\documentclass[10pt,conference]{IEEEtran}

\usepackage{graphicx}
\usepackage{amsmath}
\usepackage{amssymb}
\usepackage{booktabs}
\usepackage{threeparttable}
\usepackage{hyperref}
\usepackage{xcolor}
\usepackage{algorithm}
\usepackage{algpseudocode}

\usepackage{amsthm}
\newtheorem{theorem}{Theorem}

\begin{document}

\title{Right Labels, Wrong Reasons: Signed Feature Attribution for Multi-Label Explainability}

\newif\ifanonymous
\anonymousfalse  % camera-ready; set to \anonymoustrue for a double-anonymous review copy

\newif\ifarxiv
\arxivtrue  % IEEE camera-ready; set to \arxivtrue for the arXiv copy (personal-use notice, no page break before references)

\ifanonymous
  \author{\IEEEauthorblockN{Anonymous Author(s)}
  \IEEEauthorblockA{Anonymous Institution}}
\else
  \author{\IEEEauthorblockN{Kenton White\textsuperscript{*}}
  \IEEEauthorblockA{AC Online\\
  Algonquin College\\
  Ottawa, ON, Canada\\
  whitek@algonquincollege.com}}
\fi

\maketitle
\ifanonymous\else
  \begin{NoHyper}\renewcommand\thefootnote{*}\footnotetext{Present address: Sunshine Coast, QLD, Australia.}\end{NoHyper}
\fi
\ifarxiv
  \begin{NoHyper}\renewcommand\thefootnote{}\footnotetext{\copyright~2027 IEEE. Personal use of this material is permitted. Permission from IEEE must be obtained for all other uses, in any current or future media, including reprinting/republishing this material for advertising or promotional purposes, creating new collective works, for resale or redistribution to servers or lists, or reuse of any copyrighted component of this work in other works.}\end{NoHyper}
\fi

\begin{abstract}
When a multi-label classifier assigns the right labels, it may do so for
the wrong reasons.  Diffuse feature combinations may vary across classifier
families and lack a dominant feature.  This paper
proposes FLEX (Fisher LDA Explainer), a global surrogate that fits a
single direct multi-label Fisher LDA over the full binary label matrix,
producing signed feature attribution lists for each label in a single
coherent coordinate system.  Across the evaluated benchmarks and
classifier families, FLEX reproduces the black box's predictions with
consistently high fidelity.  Pooling covariance across labels outperforms
a per-label alternative on most datasets, with the advantage growing
largest where labels are numerous and positive examples are scarce.
FLEX's fidelity-complexity Pareto curve is monotonically increasing,
giving practitioners a predictable path to higher-fidelity explanations
as more features are used, and its in-sample fidelity matches or leads Binary Relevance
alternatives once a modest feature budget is allowed.  The magnitude of
each discriminant weight vector bounds its label's separability,
flagging labels that may be predicted correctly for the wrong reasons.  The
evaluated classifiers largely agree on which predictions to make, but
reach those decisions through different feature subsets when no single
feature dominates.
\end{abstract}

% ============================================================
\section{Introduction}
% ============================================================

Multi-label classifiers tag documents, diagnose medical scans with multiple
conditions, annotate images with multiple objects, and assign genes multiple
functional roles~\cite{zhang2014review}. They are also increasingly used with
large language models for moderation
systems~\cite{markov2023holistic}. A challenge with multi-label classifiers is that every instance can carry
several labels from a shared feature space at once.  As the number of labels
grows, each individual label's own positive examples make up a shrinking
fraction of the data~\cite{charte2015imbalance}.  With
few positive examples to constrain it, a classifier's decision for a label
can lock onto features that are predictive but weak and brittle rather than
a genuine, generalizable signal~\cite{babbar2019scarcity}. This risk
grows with every additional label competing for the same limited evidence.

Explaining a multi-label classifier's decisions is no easier.  An explainer
fit to a single label draws on that label's own limited positive examples.
Feature-importance reports built this way grow unstable across models
as those examples thin out~\cite{yuan2023background},
and the explainer reports whichever weak, brittle features the scarce
evidence happened to support~\cite{ribeiro2016lime}.  A label predicted
correctly on a weak, rather than a dominant,
feature can still be a right label, but for the wrong reason.

A reason that cannot be distinguished as genuine or spurious leaves practitioners
unable to trust any single explanation over another, however correct the
label.  FLEX (\emph{Fisher LDA Explainer}) fits a single direct multi-label
Fisher LDA over the full binary label matrix, producing signed feature
attribution lists for every label in one coherent coordinate system rather
than one independent fit per label.  The magnitude of each label's
discriminant weight vector bounds its separability.  A label whose positive
and negative groups are well separated produces a large weight vector, and
one whose groups overlap produces a small one.  Weight magnitude is
therefore a black-box diagnostic for right versus wrong reasons.

% ============================================================
\section{Related Work}
% ============================================================

Explainability for multi-label classification has received substantially
less attention than single-label XAI. Ciaperoni et al.~\cite{ciaperoni2023concise} observe that although
multi-label classification has been extensively studied, significantly
less attention has been paid to its interpretability. Panigutti et al.~\cite{panigutti2020doctorxai} propose Doctor XAI, a model-agnostic
explainer for multi-label sequential data that uses an ontology to
incorporate label correlations into local explanations.  Their work
explicitly argues that label structure should inform the explanation. 
Several extensions of LIME to the multi-label setting
have been proposed~\cite{kakogeorgiou2021evaluating}, typically by treating each label
as an independent binary classification and concatenating per-label
explanations.  The same decomposition can be instantiated with any interpretable base learner, such as decision trees~\cite{tanaka2012brct} and Lasso regression~\cite{wan2016sparse}.  Chen~\cite{chen2021shapley} extends SHAP to multi-label classification,
producing instance-level Shapley attributions per label. A recent systematic review of XAI approaches for
class differentiation~\cite{romero2026systematic} concludes that a formal framework connecting feature
contributions to class separation is still lacking in a multi-label setting.

The right labels, wrong reasons idea that multiple models with near-identical accuracy can disagree
on individual predictions is observed and formalized in predictive multiplicity~\cite{marx2020multiplicity}. The Rashomon set~\cite{fisher2019rashomon}
characterizes the full class of such near-optimal models.  Existing work
on predictive multiplicity is confined to binary classification where 
discrepancy and ambiguity are defined for single decisions, and the
optimization methods used to characterize the Rashomon boundary do not
extend directly to the multi-label setting.  

% ============================================================
\section{Method}
% ============================================================

For a multi-label dataset with $n$ samples, $m$ features, and $L$ labels,
let $Y \in \{0,1\}^{n \times L}$ be the binary label matrix ($Y_{ij}=1$
if sample $i$ carries label $j$).  Let $\mu^1_j$ and $\mu^0_j$ denote
the centroids of the positive and negative groups for label $j$.

FLEX computes a single pooled within-class covariance over all label groups,
\[
  S_w = \frac{1}{\sum_j n_j}
        \sum_{j=1}^{L} \sum_{i:\,Y_{ij}=1}
        (x_i - \mu^1_j)(x_i - \mu^1_j)^\top,
\]
where $n_j = \sum_i Y_{ij}$.  Pooling scatter across all $L$ label groups produces a better-conditioned estimate than per-label alternatives, particularly on high-cardinality datasets with rare labels where per-label $S_w$ matrices are severely rank-deficient~\cite{raninen2021coupled}.

The weight matrix $W \in \mathbb{R}^{m \times L}$ is obtained from a
single regularized Cholesky solve,
\[
  W = \bigl(S_w + \gamma I\bigr)^{-1} \Delta^\top,
  \qquad \Delta_j = \mu^1_j - \mu^0_j,
\]

Using the mean-difference $\Delta_j$ as the
right-hand side rather than the raw centroid $\mu^1_j$ gives the optimal
shared-covariance binary discriminant for each label independently
\cite{park2008multiclass}.

Column $W_{:,j}$ is the weight vector for label $j$.  Because each column
is an \emph{independent} binary discriminant, features are attributed
per-column rather than by cross-column competition.  Feature $i$ is included
in label $j$'s attribution if a two-sample $F$-test comparing feature $i$
between label $j$'s positive and negative groups is significant at level
$\alpha$ after Bonferroni correction across label $j$'s own family of $m$
features, with $\alpha = 0.05$ by default.  Included features are then
signed by $W_{i,j}$:
\begin{itemize}
  \item \textbf{Positive list of label $j$} --- if $W_{i,j} > 0$;
        the feature counts toward label~$j$.
  \item \textbf{Negative list of label $j$} --- if $W_{i,j} < 0$;
        the feature counts against label~$j$.
\end{itemize}

Because attribution is evaluated independently per column, a feature may
appear in multiple labels' positive or negative lists simultaneously, reflecting genuine co-occurrence structure in the data.  
The explainability layer receives only the raw feature matrix $X$ and the
classifier's binary predictions $\hat{Y}$; the classifier itself is not
inspected.

Algorithm~\ref{alg:flex} summarizes the complete FLEX procedure.

\begin{algorithm}[t]
\caption{FLEX: Fisher LDA Explainer for Multi-Label Classification}
\label{alg:flex}
\begin{algorithmic}[1]
\Require Feature matrix $X \in \mathbb{R}^{n \times m}$;
         black-box classifier predictions $\hat{Y} \in \{0,1\}^{n \times L}$;
         significance level $\alpha \in (0,1)$ (default $0.05$);
         top-$k$ budget per label (optional, for Pareto analysis)
\Ensure  Weight matrix $W \in \mathbb{R}^{m \times L}$;
         signed attribution lists $\mathcal{P}_j, \mathcal{N}_j$ for each label $j$

\For{$j = 1, \ldots, L$}
  \State $n_j \gets \sum_i \hat{Y}_{ij}$
  \State $\mu^1_j \gets \tfrac{1}{n_j}\sum_{i:\hat{Y}_{ij}=1} x_i$,\quad
         $\mu^0_j \gets \tfrac{1}{n - n_j}\sum_{i:\hat{Y}_{ij}=0} x_i$
  \State $\Delta_j \gets \mu^1_j - \mu^0_j$
\EndFor

\State $S_w \gets \dfrac{1}{\sum_j n_j}
        \displaystyle\sum_{j=1}^{L} \sum_{i:\,\hat{Y}_{ij}=1}
        (x_i - \mu^1_j)(x_i - \mu^1_j)^\top$ \hfill$\triangleright$\,\begin{tabular}[c]{@{}l@{}}shared\\covariance\end{tabular}

\State $\gamma \gets \max\!\bigl(\mathrm{median}(\mathrm{diag}(S_w)),\; 10^{-6}\bigr)$

\State Factorize $S_w + \gamma I = U^\top U$ \Comment{Cholesky}

\State $W \gets (S_w + \gamma I)^{-1} \Delta^\top$ \Comment{batched solve, all $L$ RHS}

\If{top-$k$ budget specified}
  \For{$j = 1, \ldots, L$}
    \State Retain top-$k$ entries of $\lvert W_{:,j} \rvert$; zero the rest
  \EndFor
\EndIf

\For{$j = 1, \ldots, L$}
  \State Compute two-sample $F$-test $p$-values $P_{i,j}$ for each feature
         $i$ (label-$j$-positive vs.\ label-$j$-negative group)
  \State Bonferroni-correct $P_{i,j}$ within label $j$'s family of $m$ features
  \State $\mathrm{Sig}_j \gets \{i : P_{i,j}^{\mathrm{corr}} < \alpha\}$
  \State $\mathcal{P}_j \gets \{i \in \mathrm{Sig}_j : W_{i,j} > 0\}$
  \State $\mathcal{N}_j \gets \{i \in \mathrm{Sig}_j : W_{i,j} < 0\}$
\EndFor

\State \Return $W$, $\{\mathcal{P}_j\}_{j=1}^L$, $\{\mathcal{N}_j\}_{j=1}^L$
\end{algorithmic}
\end{algorithm}

\subsection{Diagnostic Properties}

The weight magnitude $\lVert W_{:,j}\rVert$ is a black-box diagnostic for separability.  When label $j$'s positive and negative groups are well separated, $\lVert W_{:,j}\rVert$ is large relative to other labels.  When they overlap, it is small.  A label predicted correctly from poorly separated groups may rest on weak features, a right label for the wrong reason.

\begin{theorem}[Weight magnitude bounds separability]
\label{thm:sandwich}
Let $\Lambda = S_w + \gamma I \succ 0$, $\lVert u \rVert_{\Lambda^{-1}} = \sqrt{u^\top \Lambda^{-1} u}$, and $\lVert \cdot \rVert$ denote the Euclidean norm. For each label $j$ with $\Delta_j \neq 0$,
\begin{equation}
  \lambda_{\max}(\Lambda)^{-1/2}\,\lVert\Delta_j\rVert_{\Lambda^{-1}}
  \;\le\; \lVert W_{:,j}\rVert
  \;\le\; \lambda_{\min}(\Lambda)^{-1/2}\,\lVert\Delta_j\rVert_{\Lambda^{-1}}.
  \label{eq:sandwich}
\end{equation}
\end{theorem}

\begin{proof}
Since $W_{:,j} = \Lambda^{-1}\Delta_j$,
\[
  \lVert W_{:,j}\rVert^2 = \Delta_j^\top \Lambda^{-2} \Delta_j.
\]
Set $v = \Lambda^{-1/2}\Delta_j$, so
$\lVert v\rVert^2 = \Delta_j^\top\Lambda^{-1}\Delta_j = \lVert\Delta_j\rVert_{\Lambda^{-1}}^2$
and $\Delta_j^\top\Lambda^{-2}\Delta_j = v^\top\Lambda^{-1}v$.
The eigenvalues of $\Lambda^{-1}$ lie in
$[\lambda_{\max}(\Lambda)^{-1},\,\lambda_{\min}(\Lambda)^{-1}]$, so the
Rayleigh quotient gives
\[
  \lambda_{\max}(\Lambda)^{-1}\,\lVert\Delta_j\rVert_{\Lambda^{-1}}^2
  \;\le\; \lVert W_{:,j}\rVert^2
  \;\le\; \lambda_{\min}(\Lambda)^{-1}\,\lVert\Delta_j\rVert_{\Lambda^{-1}}^2.
\]
Taking square roots gives~\eqref{eq:sandwich}.
\end{proof}

Since $\Lambda$ is shared across all labels, $\lVert W_{:,j}\rVert$ is a
cross-label proxy for separability bounded
up to $\sqrt{\kappa(\Lambda)}$, where $\kappa(\Lambda) = \lambda_{\max}(\Lambda)/\lambda_{\min}(\Lambda)$ is the condition number of $\Lambda$.

% ============================================================
\section{Experimental Design}
% ============================================================

Evaluation uses five standard multi-label benchmarks drawn from the survey
of Madjarov et al.~\cite{madjarov2012extensive}, spanning genomics, music,
image, and email domains and covering a wide range of label cardinalities
(mean 1.07--4.24 labels per sample) and label-space sizes (6--53 labels).

Six classifiers spanning the main paradigms in multi-label learning are trained on a 50/50 split using iterative stratification~\cite{sechidis2011stratification} to preserve rare-label distribution across both halves:

\begin{itemize}
  \item \textbf{BR-LR:} Binary Relevance with Logistic Regression.  The linear
    classifier provides a sanity check for when the attribution rules are essentially
    recovering a linear decision boundary.
  \item \textbf{BR-RF:} Binary Relevance with Random Forest (200 trees).  The
    canonical nonlinear baseline. Its per-label independence makes it
    the most interpretable black box.
  \item \textbf{BR-XGB:} Binary Relevance with XGBoost (200 rounds).  A
    gradient-boosted alternative to BR-RF. Higher predictive power
    at the cost of a more complex decision boundary.
  \item \textbf{ECC:} Ensemble of 10 Classifier Chains~\cite{read2011classifier}
    with Random Forest base learners and random label orderings.
    Predictions are averaged and thresholded at~0.5.  ECC explicitly
    models label co-occurrence.
  \item \textbf{ML-kNN:} Multi-label $k$-nearest neighbours~\cite{zhang2007ml}
    ($k=10$).  A non-parametric baseline whose decision boundary has no
    obvious linear structure.
  \item \textbf{RAkELo:} Random $k$-labelsets (overlapping)~\cite{tsoumakas2011random}
    with Random Forest base learners.  Samples $2L$ overlapping labelsets of
    size~3 and trains a multi-class classifier per subset, capturing
    within-subset label interactions.
\end{itemize}

The FLEX explainability layer operates identically across all classifiers.
It receives only the raw feature matrix $X$ and the binary prediction
matrix $\hat{Y}$, with no access to classifier internals.  

Classifier performance is measured by F1 averaged across labels and
Hamming loss against ground truth.  Explainability fidelity is
measured three ways against the \emph{classifier's own predictions}~$\hat{Y}$: 
precision, recall and averaged ROC-AUC of the raw continuous FLEX scores against
$\hat{Y}$.  All reported values pool predictions across all labels before scoring.

% ============================================================
\section{Experimental Results}
% ============================================================

\subsection{Classifier Performance and Fidelity}

Table~\ref{tab:classifier} reports F1 and Hamming loss for all six classifiers; Table~\ref{tab:fidelity} reports FLEX precision, recall, and AUC against each classifier's predictions.\footnote{\url{https://github.com/KentonWhite/multi-label-explainability-replication}} Three patterns in fidelity emerge.

\begin{table}[t]
\centering
\caption{Classifier performance on the test set (50/50 iterative-stratification split,
  mean\,$\pm$\,std over 10 seeds).}
\label{tab:classifier}
\begin{threeparttable}
\scriptsize
\setlength{\tabcolsep}{3pt}
\begin{tabular}{lccccc}
\toprule
Classifier & Yeast & Emotions & Scene & Genbase\tnote{a} & Enron \\
\midrule
\multicolumn{6}{l}{\textit{F1}} \\
BR-LR   & $.628{\pm}.004$ & $.630{\pm}.016$ & $.687{\pm}.009$ & $.981{\pm}.004$ & $.543{\pm}.009$ \\
BR-RF   & $.634{\pm}.004$ & $.670{\pm}.010$ & $.693{\pm}.009$ & $.992{\pm}.003$ & $.553{\pm}.008$ \\
BR-XGB  & $.640{\pm}.004$ & $.660{\pm}.018$ & $.745{\pm}.009$ & $.975{\pm}.004$ & $.562{\pm}.006$ \\
ECC     & $.644{\pm}.005$ & $.676{\pm}.013$ & $.704{\pm}.012$ & $.993{\pm}.002$ & $.558{\pm}.010$ \\
ML-kNN  & $.631{\pm}.012$ & $.649{\pm}.013$ & $.722{\pm}.010$ & $.956{\pm}.013$ & $.431{\pm}.022$ \\
RAkELo  & $.646{\pm}.006$ & $.687{\pm}.016$ & $.744{\pm}.010$ & $.993{\pm}.003$ & $.552{\pm}.010$ \\
\midrule
\multicolumn{6}{l}{\textit{Hamming Loss}} \\
BR-LR   & $.202{\pm}.002$ & $.212{\pm}.007$ & $.103{\pm}.003$ & $.003{\pm}.001$ & $.050{\pm}.001$ \\
BR-RF   & $.193{\pm}.002$ & $.187{\pm}.005$ & $.089{\pm}.002$ & $.001{\pm}.000$ & $.047{\pm}.001$ \\
BR-XGB  & $.200{\pm}.002$ & $.200{\pm}.011$ & $.082{\pm}.002$ & $.004{\pm}.001$ & $.050{\pm}.001$ \\
ECC     & $.192{\pm}.002$ & $.187{\pm}.007$ & $.087{\pm}.003$ & $.001{\pm}.000$ & $.047{\pm}.001$ \\
ML-kNN  & $.206{\pm}.002$ & $.208{\pm}.007$ & $.093{\pm}.003$ & $.006{\pm}.002$ & $.058{\pm}.002$ \\
RAkELo  & $.193{\pm}.003$ & $.185{\pm}.009$ & $.080{\pm}.003$ & $.001{\pm}.000$ & $.047{\pm}.001$ \\
\bottomrule
\end{tabular}
\begin{tablenotes}
  \scriptsize
  \item[a] 11 of 27 Genbase labels have fewer than 10 positives and are excluded; 16 labels retained.
\end{tablenotes}
\end{threeparttable}
\end{table}

\textit{AUC is uniformly high.}  ROC-AUC meets or exceeds
$0.90$ across all dataset--classifier pairs, and exceeds $0.96$ in 25 of 30
cases.  The two lowest cases --- ML-kNN on Yeast ($0.901\pm0.009$) and
BR-XGB on Yeast ($0.911\pm0.003$) --- reflect the greater non-linearity of
these classifiers.

\textit{Recall is consistently high.}  Across all datasets, recall is at least $0.77$ and is at or near~$1.00$ for Genbase,
Scene, and Enron.  The FLEX attribution rules recover the large majority of
positive predictions made by the target classifier.

\textit{Precision varies with the classifier and the data.}
BR-LR achieves the highest precision on Yeast ($0.876\pm0.008$).
RF- and ensemble-based methods yield modestly lower Yeast precision
(BR-RF: $0.819\pm0.009$, ECC: $0.829\pm0.013$).  Genbase precision varies widely ($0.245$--$0.579$)
despite perfect recall because the binary Prosite feature matrix produces a
near-singular within-class covariance.  Scene shows lower precision ($0.55$--$0.65$) with
near-perfect recall, reflecting its near-single-label structure.

\begin{table}[t]
\centering
\caption{Explainability fidelity: FLEX vs.\ the classifier's predictions
  $\hat{Y}$ on the test set (mean\,$\pm$\,std, 10 seeds, 50/50 split).
  FLEX is fit and scored on the same $(X_{\text{test}}, \hat{Y}_{\text{test}})$ (in-sample).
  Labels with fewer than 5 positive test-set predictions are excluded.}
\label{tab:fidelity}
\begin{threeparttable}
\scriptsize
\setlength{\tabcolsep}{3pt}
\begin{tabular}{lccccc}
\toprule
Classifier & Yeast & Emotions & Scene & Genbase\tnote{b} & Enron \\
\midrule
\multicolumn{6}{l}{\textit{Precision}} \\
BR-LR   & $.876{\pm}.008$ & $.737{\pm}.015$ & $.648{\pm}.015$ & $.579{\pm}.148$ & $.887{\pm}.010$ \\
BR-RF   & $.819{\pm}.009$ & $.730{\pm}.012$ & $.553{\pm}.012$ & $.364{\pm}.074$ & $.889{\pm}.019$ \\
BR-XGB  & $.685{\pm}.010$ & $.697{\pm}.017$ & $.576{\pm}.013$ & $.378{\pm}.128$ & $.887{\pm}.005$ \\
ECC     & $.829{\pm}.013$ & $.739{\pm}.016$ & $.558{\pm}.010$ & $.365{\pm}.074$ & $.896{\pm}.022$ \\
ML-kNN  & $.683{\pm}.023$ & $.735{\pm}.027$ & $.585{\pm}.013$ & $.245{\pm}.090$ & $.853{\pm}.031$ \\
RAkELo  & $.818{\pm}.037$ & $.758{\pm}.016$ & $.585{\pm}.009$ & $.365{\pm}.074$ & $.892{\pm}.022$ \\
\midrule
\multicolumn{6}{l}{\textit{Recall}} \\
BR-LR   & $.934{\pm}.011$ & $.934{\pm}.007$ & $.983{\pm}.003$ & $1.00{\pm}.00$ & $.941{\pm}.003$ \\
BR-RF   & $.937{\pm}.012$ & $.942{\pm}.010$ & $.979{\pm}.006$ & $1.00{\pm}.00$ & $.939{\pm}.006$ \\
BR-XGB  & $.801{\pm}.008$ & $.901{\pm}.010$ & $.957{\pm}.006$ & $1.00{\pm}.00$ & $.932{\pm}.005$ \\
ECC     & $.933{\pm}.015$ & $.939{\pm}.011$ & $.978{\pm}.007$ & $1.00{\pm}.00$ & $.941{\pm}.004$ \\
ML-kNN  & $.778{\pm}.016$ & $.911{\pm}.015$ & $.944{\pm}.007$ & $1.00{\pm}.00$ & $.919{\pm}.011$ \\
RAkELo  & $.929{\pm}.021$ & $.925{\pm}.009$ & $.971{\pm}.005$ & $1.00{\pm}.00$ & $.941{\pm}.004$ \\
\midrule
\multicolumn{6}{l}{\textit{AUC}} \\
BR-LR   & $.979{\pm}.003$ & $.970{\pm}.004$ & $.988{\pm}.001$ & $.986{\pm}.003$ & $.995{\pm}.001$ \\
BR-RF   & $.979{\pm}.004$ & $.969{\pm}.004$ & $.984{\pm}.002$ & $.981{\pm}.006$ & $.994{\pm}.001$ \\
BR-XGB  & $.911{\pm}.003$ & $.947{\pm}.004$ & $.974{\pm}.002$ & $.976{\pm}.009$ & $.994{\pm}.001$ \\
ECC     & $.978{\pm}.004$ & $.970{\pm}.004$ & $.984{\pm}.002$ & $.981{\pm}.006$ & $.994{\pm}.002$ \\
ML-kNN  & $.901{\pm}.009$ & $.956{\pm}.007$ & $.969{\pm}.002$ & $.956{\pm}.021$ & $.994{\pm}.001$ \\
RAkELo  & $.977{\pm}.007$ & $.964{\pm}.005$ & $.980{\pm}.002$ & $.981{\pm}.006$ & $.994{\pm}.002$ \\
\bottomrule
\end{tabular}
\begin{tablenotes}
  \scriptsize
  \item[b] Genbase fidelity computed on 16 retained labels (see Table~\ref{tab:classifier}, note~a).
\end{tablenotes}
\end{threeparttable}
\end{table}

\subsection{Effect of Covariance Pooling}

Table~\ref{tab:ablation} compares FLEX (shared pooled $S_w$) against
BR-LDA (per-label $S_w$)
using BR-RF as the black-box target.  The two methods are identical in
every respect except for the covariance scope: same Fisher LDA family,
same regularization strategy ($\gamma = \mathrm{median}(\mathrm{diag}(S_w))$,
floored at $10^{-6}$), same RHS construction, same feature attribution
protocol.  Any fidelity difference is attributable solely to covariance
pooling.

\begin{table}[t]
\centering
\caption{Effect of covariance pooling: FLEX (shared $S_w$) vs.\ BR-LDA (per-label $S_w$) on BR-RF predictions.
  Both are fit and scored on the same $(X_{\text{test}}, \hat{Y}_{\text{test}})$ (in-sample).
  P~=~Precision, R~=~Recall, AUC~=~ROC-AUC
  (mean\,$\pm$\,std, 10 seeds).}
\label{tab:ablation}
\begin{threeparttable}
\scriptsize
\setlength{\tabcolsep}{3pt}
\resizebox{\linewidth}{!}{%
\begin{tabular}{lccc ccc}
\toprule
 & \multicolumn{3}{c}{Shared (FLEX)} & \multicolumn{3}{c}{BR-LDA} \\
\cmidrule(lr){2-4}\cmidrule(lr){5-7}
Dataset & P & R & AUC & P & R & AUC \\
\midrule
Yeast     & $.819{\pm}.009$ & $.937{\pm}.012$ & $.979{\pm}.004$ & $.732{\pm}.022$ & $.928{\pm}.088$ & $.963{\pm}.011$ \\
Emotions  & $.736{\pm}.021$ & $.946{\pm}.008$ & $.969{\pm}.005$ & $.637{\pm}.023$ & $.978{\pm}.005$ & $.960{\pm}.006$ \\
Scene     & $.554{\pm}.007$ & $.978{\pm}.005$ & $.984{\pm}.001$ & $.409{\pm}.007$ & $.998{\pm}.001$ & $.963{\pm}.005$ \\
Genbase\tnote{c} & $.334{\pm}.041$ & $1.00{\pm}.00$ & $.980{\pm}.003$ & $.967{\pm}.015$ & $1.00{\pm}.00$ & $.999{\pm}.001$ \\
Enron     & $.889{\pm}.011$ & $.937{\pm}.004$ & $.994{\pm}.001$ & $.323{\pm}.032$ & $.980{\pm}.003$ & $.814{\pm}.033$ \\
\bottomrule
\end{tabular}%
}
\vspace{2pt}
\noindent\parbox{\linewidth}{\scriptsize\textsuperscript{c}~Genbase is the
exception: per-label $S_w$ wins on AUC (0.999 vs.\ 0.980).  Each Genbase
label corresponds to a distinct Prosite functional class, so the
label-specific covariance captures signal that pooling obscures.}
\end{threeparttable}
\end{table}

Shared pooling wins on AUC in four of five datasets, with the margin
growing with label cardinality.  On Enron (53~labels), the gap reaches
$\Delta\mathrm{AUC}=+0.180$: the per-label $S_w$ matrices are on average
$184{,}000\times$ worse conditioned than the shared $S_w$ (per-label mean
condition number $6.32 \times 10^7$ vs.\ shared $3.44 \times 10^2$).
Table~\ref{tab:condnum} reports condition numbers across both conditions.

\begin{table}[t]
\caption{Condition numbers $\kappa(S_w)$ across two covariance strategies
  (mean\,$\pm$\,std, 10 seeds).
  Shared pools scatter across all labels; BR-LDA uses per-label $S_{w,j}$.}
\label{tab:condnum}
\scriptsize
\setlength{\tabcolsep}{5pt}
\centering
\begin{tabular}{lcc}
\toprule
Dataset & Shared & BR-LDA (mean) \\
\midrule
Yeast    & $11.8{\pm}0.2$          & $25.0{\pm}5.6$          \\
Emotions & $12.5{\pm}0.5$          & $20.5{\pm}1.3$          \\
Scene    & $27.9{\pm}1.0$          & $43.1{\pm}1.4$          \\
Genbase  & $5.72{\pm}1.46\times10^{6}$ & $3.54{\pm}0.28\times10^{7}$ \\
Enron    & $344{\pm}20$            & $6.32{\pm}1.30\times10^{7}$ \\
\bottomrule
\end{tabular}
\end{table}

The shared $S_w$ is well-conditioned because it pools scatter across all
label groups, accumulating far more data than any single label can provide.
On Enron each of the 53 labels has on average only 64 positives in a
1001-dimensional feature space ($n_j \ll m$); the per-label $S_w$ is
severely rank-deficient before regularization.  Pooling across all labels
effectively increases the sample count driving the covariance estimate by
a factor proportional to $L$, which is precisely the regime where
shared-covariance LDA is known to dominate per-class
estimation~\cite{raninen2021coupled}.

\subsection{Fidelity--Complexity Pareto Curves}

\begin{figure}[t]
  \centering
  \includegraphics[width=\linewidth]{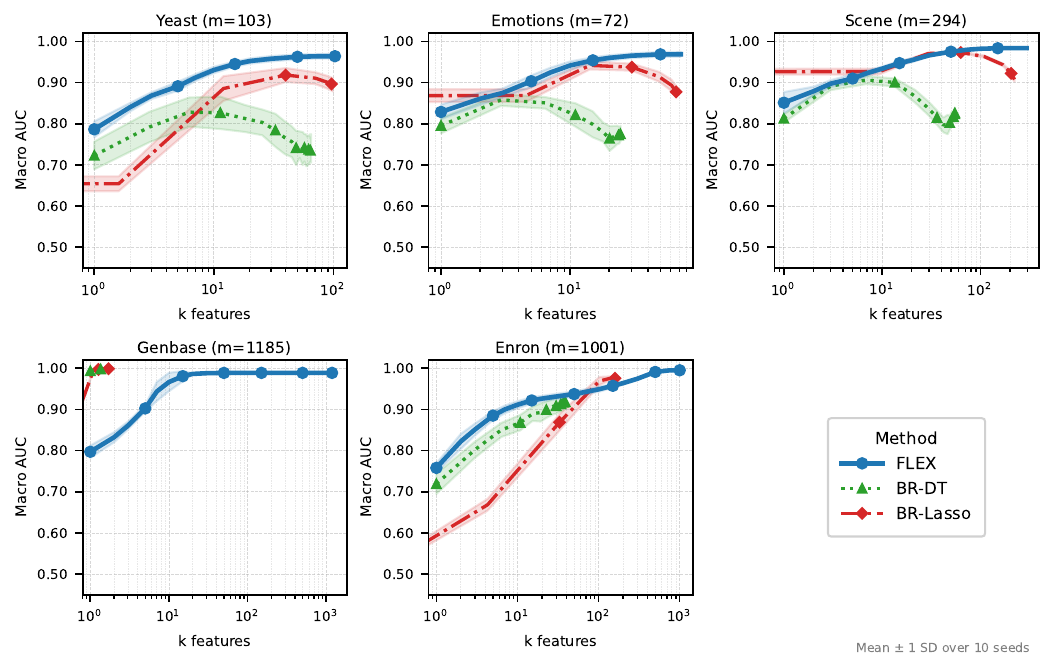}
  \caption{Fidelity--complexity Pareto curves.  $x$-axis: $k$ features per
    label (log scale).  $y$-axis: AUC of surrogate vs.\ black-box
    predictions on $(X_{\text{test}},\hat{Y}_{\text{test}})$.
    FLEX (closed-form) fits on the full test pool;
    BR-DT and BR-Lasso train on $(X_{\text{train}},\hat{Y}_{\text{train}})$.}
  \label{fig:pareto}
\end{figure}

A practical explanation method must be robust to the budget choice $k$. Not knowing the right budget in advance should not be a prerequisite for fidelity.
Figure~\ref{fig:pareto} reveals a sharp asymmetry in budget robustness across
methods.

Figure~\ref{fig:pareto} shows AUC as a function of feature
complexity $k$ for FLEX, BR-DT, and BR-Lasso across all five datasets.
FLEX is closed-form and fits on the full $(X_{\text{test}},\hat{Y}_{\text{test}})$;
BR-DT and BR-Lasso train on $(X_{\text{train}},\hat{Y}_{\text{train}})$ and are
scored on $(X_{\text{test}},\hat{Y}_{\text{test}})$.  FLEX fidelity is therefore
in-sample and BR-DT and BR-Lasso fidelity is out-of-sample, so FLEX's curve
reports fidelity only on the instances it was fit to.  For FLEX,
$k$ is controlled by retaining only the top-$k$ features per label by weight
magnitude $|w_j[i]|$ and recomputing scores on the reduced support. For BR-DT,
$k$ is the mean number of unique features used across labels at a given tree
depth. For BR-Lasso, $k$ is the mean number of non-zero coefficients at a given
regularization strength $C$.  Consequently, BR-DT and BR-Lasso curves
terminate before $k=m$. Tree depth is bounded by the number of splits that
improve training-set accuracy, and the Lasso path ends when all coefficients
become non-zero, both of which occur well below $m$ on most datasets.
FLEX by contrast sweeps the full range from $k=1$ to $k=m$ by
construction.

Three patterns are consistent across datasets.  First, FLEX is
\emph{budget-robust}: its curve is monotonically increasing, so any
choice of $k$ yields reliable fidelity, and at $k=m$ FLEX achieves the
highest AUC of any method on Yeast ($0.964$), Emotions ($0.969$), Scene
($0.984$), and Enron ($0.995$).  The Fisher discriminant accumulates
discriminative evidence from each additional feature,
making the budget choice inconsequential for fidelity.

Second, BR-DT is \emph{budget-brittle}: it peaks early and degrades with
depth.  On Emotions BR-DT reaches AUC$\,{=}\,0.858$ at $k \approx 3$
features but falls to $0.764$ at $k \approx 20$. On Yeast it peaks at
$0.828$ ($k \approx 6$) and degrades to $0.741$ by $k \approx 50$.  A
practitioner who sets the budget beyond the shallow peak will receive a
lower-fidelity explanation than one who stopped earlier.  Decision trees
overfit the training-set labels beyond shallow splits.

Third, BR-Lasso is \emph{erratic at low $k$}: L1 regularization either
selects the most predictive feature or collapses to near-chance depending on
the regularization path, so fidelity at small budgets is sensitive to an
arbitrary regularization choice~\cite{zou2005elastic}.

At low $k$ ($k \leq 5$), all methods are broadly competitive. BR-DT holds a
small advantage on Scene and Genbase where a single highly-discriminative
feature dominates.  Genbase is the consistent exception (BR-DT reaches
AUC $\geq 0.993$ at $k=1$). Its near-orthogonal binary Prosite features
mean a single feature is already near-sufficient, and the shared $S_w$ must
accumulate many features before recovering.  On all other datasets FLEX
matches or leads the Pareto frontier from $k \approx 20$ onwards.  The
practical consequence is that BR-based explainers require careful budget
selection to avoid low-fidelity explanations, while FLEX's rising
curve makes the budget choice inconsequential.

% ============================================================
\section{Discussion}
% ============================================================

\subsection{Comparison with Binary Relevance}

The experiments highlight two important differences between FLEX and Binary Relevance.  First, the shared within-class covariance
$S_w$ is consistently better conditioned than per-label estimates, and the
benefit is largest precisely where it matters: high-cardinality datasets with
rare labels.  On Enron the per-label condition number is five orders of
magnitude worse than the shared estimate, and the corresponding AUC gap
($+0.180$) is the largest.  This is precisely
the regime where pooling is theoretically motivated~\cite{friedman1989rda}, and
the results confirm that the theory transfers to practice.

Second, FLEX is robust to the feature budget $k$: its Pareto curve
increases monotonically on every dataset without tuning.
Features are added in order of discriminative magnitude, rather than by a
greedy split or a regularization path, and fidelity rises as features are added.  BR-DT and BR-Lasso lack this property.
Tree overfitting degrades fidelity beyond the shallow peak, and Lasso's L1
path is erratic at the low-$k$ end.  The budget robustness of FLEX means
practitioners receive a predictable, high-fidelity explanation at any
complexity level, without needing to tune the budget to avoid explanation
degradation.

\textit{Yeast case study.}
The covariance-pooling gap is easiest to see at the per-label level.
Yeast (mean 4.24 labels per gene, 14 labels total) is the dataset where the
gap between shared and per-label covariance is most instructive.  The per-label
breakdown of the covariance pooling shows that shared pooling raises
precision uniformly across all label groups, with the largest absolute gains on
the dominant labels (Class~12, Class~13) where both methods already achieve
near-perfect AUC.  Rare labels (Class~6--8) show modest AUC parity but shared
pooling still shows higher precision (e.g.\ Class~8: shared
P$=.114$ vs.\ BR-LDA P$=.058$), consistent with the pooled estimate being more
stable when few positives are available.  The Pareto analysis further reveals
a similar two-tier structure. Class~12 and Class~13 achieve P$=$1.000,
R$\geq$0.993 at full feature budget. These dominant labels are well-separated
in feature space and the linear rules are essentially exact. But rare
labels (Class~6--8) achieve strong recall (0.82--0.92) with low precision
(0.11--0.27).  This precision gap for rare labels means FLEX predicts those
labels on more instances than the classifier does.  A likely cause is that FLEX
fires when the features that count against them are low or absent, a
consequence of the shared $S_w$ moving the covariance estimate toward the
majority labels.

\subsection{Weight Concentration as a Diagnostic Signal}

Beyond these structural properties, FLEX's weights may also signal
explanation reliability.  This connects FLEX to the broader disagreement problem in explainable
ML~\cite{krishna2022disagreement}, where different explanation methods applied
to the same model routinely produce conflicting attributions.  When a
label lacks a dominant feature, different classifier families exploit different
arbitrary feature subsets, leading to predictive
multiplicity~\cite{marx2020multiplicity,fisher2019rashomon} that propagates
directly into explanation instability~\cite{alvarez2018robustness}.
To verify that this instability reflects genuine signal ambiguity rather than
prediction-level disagreement, pairwise discrepancy~\cite{kuncheva2003measures} is measured
across all 15 classifier pairs on each dataset.  Mean pairwise discrepancy ranges
from $0.4\%$ (Genbase) to $10.7\%$ (Emotions), showing that the classifiers largely agree on
which instances to label positive.  The observed multiplicity is therefore in
the \emph{explanation}. Different classifiers reach the same decisions via
different feature subsets rather than in the predictions themselves.

In the benchmark datasets one feature can dominate label $j$, which shows in
the shape of $W_{:,j}$ rather than its magnitude.  The norm bounded by
Theorem~\ref{thm:sandwich} scales with the mean shift $\Delta_j$, which varies
across labels regardless of whether one feature dominates.
\emph{Concentration} removes this scale by dividing the largest weight by the
mean weight, $c_j = \max_i \lvert W_{i,j}\rvert / \mathrm{mean}_i \lvert W_{i,j}\rvert$.
Concentration and $\lVert W_{:,j}\rVert$ can therefore vary independently,
and the link between concentration and disagreement is tested empirically
below.

Testing whether concentration predicts disagreement requires a matching
per-pair disagreement measure.  For each classifier pair, \emph{disagreement} is defined using their own internal model labels.
Genbase and Enron have binary features (Prosite
motif presence; vocabulary presence), where a real signal can plausibly live
in a single feature. For
these, \emph{disagreement} is measured as whether the two classifiers pick the
same single top feature (top-1 match).  Yeast, Emotions, and Scene have
continuous features, where a coherent signal is more plausibly spread across
several correlated values than concentrated in one. Here a raw index match
can't be used reliably.  It would score two classifiers as disagreeing even
when they pick different but highly correlated features.  Instead, for
Yeast and Scene, features are clustered by
pairwise correlation ($\lvert r \rvert>0.7$ on $X_{\mathrm{test}}$) and
count agreement at the cluster level. Emotions' features are not
meaningfully redundant in this way, so the full weight-vector Spearman
correlation between classifiers' internal attributions is used directly.  In every
case, disagreement is averaged over all fifteen classifier pairs
across ten seeds.

\begin{table}[t]
\caption{Spearman correlation between FLEX weight concentration and
  cross-classifier disagreement among each classifier's own native internal
  feature attributions.}
\label{tab:concentration}
\scriptsize
\centering
\begin{tabular}{llccc}
\toprule
Dataset & Measure & $n$ (labels) & $\rho$ & $p$ \\
\midrule
Enron    & Top-1 match            & 32 & $-0.568$ & $0.001$ \\
Genbase  & Top-1 match            & 16 & $-0.545$ & $0.029$ \\
Emotions & Full-vector Spearman   & 6  & $-0.486$ & $0.329$ \\
Yeast    & Cluster-aware top-1    & 12 & $-0.193$ & $0.549$ \\
Scene    & Cluster-aware top-1    & 6  & $-0.086$ & $0.872$ \\
\bottomrule
\end{tabular}
\end{table}

Table~\ref{tab:concentration} summarizes the correlation between the FLEX concentration and the classifier disagreement.
All five datasets show the predicted direction: larger weights signal smaller disagreement and smaller weights signal larger disagreement. Only Enron and Genbase, the two binary-feature
datasets, reach significance of $p < 0.05$.  On the other datasets, the qualitative pattern
tracks how directly a feature measures a physical quantity. Emotions' spectral
descriptors are the most direct measurements and have the most
significance among the continuous datasets. Yeast's expression measurements
and Scene's spatial color-moment summaries are less direct and show a
weaker relationship.

\textit{Emotions case study.}
The Emotions dataset provides a clear contrastive example.
For \texttt{quiet.still}, the FLEX weight vector is strongly peaked: the
top feature (\texttt{Std\_Acc\_Mn\_MFCC\_1}, cross-segment variability of the
second mel-frequency cepstral coefficient) accounts for $6.5{\pm}1.0\%$ of
total absolute weight, with a top-to-mean ratio of $4.7{\pm}0.7\times$
averaged across all six classifiers and ten seeds.  When this FLEX
explanation is validated against model internals, the Spearman rank correlation
between the BR-RF internal feature importances and the FLEX weights reaches
$\rho = 0.444{\pm}0.078$ averaged across ten seeds --- the highest of any
label or classifier combination in the study --- with
\texttt{Std\_Acc\_Mn\_MFCC\_1} ranked first by both methods simultaneously.
ML-kNN internal attributions independently recover the same three features
(\texttt{Std\_Acc\_Mn\_MFCC\_0/1/4}) that FLEX identifies, in a different
order.  

The \texttt{happy.pleased} label presents the opposite case.  The top FLEX
weight accounts for only $4.7{\pm}0.9\%$ of total absolute weight with a
top-to-mean ratio of $3.4{\pm}0.6\times$ --- both the lowest values in the
dataset.  The number of predicted positives varies from 26 (BR-RF) to 58
(ML-kNN) across classifier families: the classifiers do not agree on which
examples \emph{are} \texttt{happy.pleased}, a direct instance of predictive
multiplicity~\cite{marx2020multiplicity}.  Validation against model internals
yields no agreement across methods, and FLEX explanation disagreement
between classifier families is the highest of any Emotions label.  These
classifiers achieve non-trivial accuracy on \texttt{happy.pleased} but do so
through diffuse, unstable feature combinations rather than a coherent
discriminative strategy.  The weak FLEX magnitude is a direct indicator of
this instability.

\textit{Enron case study.}
The same pattern appears in Enron.  For label \texttt{C.C6}, a single
vocabulary feature dominates the FLEX explanation in every classifier
family across all ten seeds, with a top-to-mean weight ratio of
$10.4{\pm}3.7\times$.  Validation against BR-LR coefficients, BR-RF feature
importances, and ML-kNN neighbourhood attributions all independently rank this
same feature first. For label \texttt{B.B4} the top-to-mean ratio falls to $5.7{\pm}1.0\times$, and
the top-20 most important features for BR-LR and ML-kNN share zero overlap
across ten seeds.  Model internals and FLEX weights disagree with each other
and across classifier families, because weight is spread across more than 600
vocabulary features with no single dominant feature.

Taken together, these results show that FLEX weight concentration is a
black-box signal for whether a classifier's predictions rest on a dominant
feature.  When one feature dominates, the classifiers' own explanations tend
to agree with FLEX and with each other.  When the weights are spread out, they
tend to disagree.  No access to model internals is required to make this
assessment.

\subsection{Limitations}

FLEX approximates the black box with one linear surrogate and one covariance
estimate.  Two conditions break that approximation.  First, when the
black-box classifier relies on highly non-linear interactions (XGB and ML-kNN
on Yeast are the clearest cases, with AUC of $0.911$ and $0.901$
respectively), the linear surrogate cannot closely approximate the non-linear
decision surface, and both precision and AUC fall.  Second, when the pooled
covariance $S_w$ is ill-conditioned (Genbase is the clearest case, where
near-orthogonal binary features drive up the condition number), fidelity
degrades even though the underlying decision boundary is comparatively
simple.  In both cases the diagnostic in the preceding subsection still
holds.  Low fidelity shows up as reduced concentration and precision rather
than a silent, confident wrong explanation.

% ============================================================
\section{Conclusion}
% ============================================================

Multi-label classifiers can achieve high accuracy through weak feature
combinations that vary across classifier families.  This paper has
presented FLEX, a global
surrogate that addresses both sides of this problem: it fits a single direct
Fisher LDA over the full binary label matrix, producing signed per-label
attribution vectors in a single coherent coordinate system, and its weight
concentration provides a black-box diagnostic for whether a
classifier's predictions rest on a dominant feature.

On five benchmark datasets against six classifiers, FLEX achieves 
AUC above 0.92 in all but the most non-linear configurations, matches or leads Binary Relevance
alternatives on in-sample fidelity from $k \approx 20$ features onwards, and
outperforms Binary Relevance decomposition by up to $+0.180$ AUC on
high-cardinality datasets.  The monotonically increasing Pareto curve gives practitioners a
predictable path from concise explanations at low $k$ to high-fidelity
reproductions at high $k$.  When one feature dominates a label's
weights, classifiers tend to agree on their explanations.  When the weights
are spread out, they tend to disagree.  The right labels, but
reached through wrong reasons.

\subsubsection*{Acknowledgments}
Claude~\cite{anthropic2025claude} (Anthropic) was used to automate the execution of benchmark
experiments, draft structured research notes (which informed early drafts of the paper), format and typeset the \LaTeX{}
manuscript, and produce the example implementation in the replication notebook.
All experimental designs, hypotheses, interpretations, and
conclusions are the authors' own.

\ifarxiv\else\clearpage\fi
\bibliographystyle{IEEEtran}
\bibliography{references}

\end{document}